\pdfoutput=1 
\documentclass[11pt]{article}

\usepackage[font=libertinus, citestyle=numeric]{kurbanlab}

\DeclareAffiliation{hbku}{%
  College of Science and Engineering, Hamad Bin Khalifa University, Doha, Qatar}

\DeclareAffiliation{tamu}{%
  Department of Computer and Electrical Engineering,
  Texas A\&M University, College Station, TX, USA}

\DeclareAffiliation{iub}{%
  Luddy School of Informatics, Computing, and Engineering,
  Indiana University Bloomington, Bloomington, IN, USA}

\usepackage{colortbl}
\usepackage{threeparttable}
\usepackage{makecell}
\usepackage{longtable}
\usepackage{pifont}
\usepackage{float}
\usepackage{bibunits}
\defaultbibliography{references}
\defaultbibliographystyle{unsrtnat}
\usetikzlibrary{arrows.meta,positioning,calc,shapes.geometric,backgrounds,fit}

\definecolor{hlcol}{RGB}{232,239,248}
\definecolor{negcol}{RGB}{251,236,229}
\definecolor{boxblue}{RGB}{225,235,248}
\definecolor{boxedge}{RGB}{70,110,160}
\definecolor{accent}{RGB}{40,90,150}
\definecolor{warm}{RGB}{200,90,40}
\definecolor{good}{RGB}{40,130,80}
\definecolor{rowhi}{RGB}{226,240,232}
\definecolor{headrow}{RGB}{225,235,248}
\definecolor{boxbg}{RGB}{238,244,251}

\newcommand{\headrowc}[1]{\multicolumn{1}{c}{\bfseries #1}}
\newcommand{\yes}{\textcolor{good}{\ding{51}}}
\newcommand{\no}{\textcolor{warm!85!black}{\ding{55}}}

\newcommand{\method}{\textsc{Pgs}}
\newcommand{\methodlong}{Perturbation-Grounded Selection}

\title{It's the Decoding Format, Not the Perturbation: Auditing Consistency-Based Selection for Vision-Language Test-Time Scaling}
\Subtitle{A format-matched control shows that apparent PGS gains over majority voting are largely a CoT$\to$short effect}
\RunningTitle{Decoding format, not perturbation, for VLM test-time scaling}

\Author[]{Puzhuo Zheng}{hbku}
\Author[corresponding=hkurban@hbku.edu.qa]{Hasan Kurban}{hbku}

\CodeURL{https://github.com/KurbanIntelligenceLab/PGS-Code-Audit-Tooling.git}   
\Venue{Preprint}

\begin{document}
\maketitle

\begin{bibunit}[unsrtnat]

\begin{abstract}
Test-time scaling lifts large language model reasoning by sampling many candidate
solutions and selecting among them, yet the same recipe transfers poorly to
vision-language models (VLMs): recent work shows that simple majority voting beats
selection methods built on the model's own self-verification, apparently because at
the selection layer an image-grounded answer and a confident guess from the language
prior look the same. A natural fix is to make the selection signal one that cannot be
computed without the image. We study \methodlong{} (\method{}), a label-free,
training-free rule that scores each candidate by whether the model re-derives it under
label-preserving perturbations of the input (cropping, background masking, mild
photometric or geometric jitter); \method{} recovers majority voting when the
perturbation set is empty. The decisive question is not whether \method{} beats
chain-of-thought-only majority voting, but whether the perturbation term adds anything
once decoding format and budget are controlled. We therefore introduce a
\emph{format-matched} control (MatchedCtrl): the same short, no-CoT draws spent on the
\emph{original} image. Across TextVQA, MATH-Vision, MMMU, and ViLP, with a Qwen headline (three-seed means)
and LLaVA-OneVision coverage in matched-budget selector tables, \method{} appears to beat
plain majority voting by up to $+31.8$ points on TextVQA (Qwen), but MatchedCtrl tracks or exceeds
\method{} within noise on every benchmark, including the vision-required ViLP; no Qwen
category shows a significant gain over this control. The preserve/destroy stability gap is real and image-dependent (up to
$+0.48$), yet does not predict per-instance wins. The result is negative and diagnostic:
perturbation consistency is at best a partial diagnostic of visual dependence and, on its
own, not a usable selection signal once format is controlled; gains reported against
CoT-only majority voting overstate such methods. We release code and audit tooling on \url{https://github.com/KurbanIntelligenceLab/PGS-Code-Audit-Tooling.git}.
\end{abstract}

\printkeywords



\begin{center}
\resizebox{0.99\textwidth}{!}{%
\begin{tikzpicture}[
  font=\sffamily\small,
  >={Latex[length=2.2mm]},
  imgbox/.style={draw=boxedge,line width=0.5pt,inner sep=0.4pt,fill=boxblue!25},
  card/.style={rounded corners=2pt,draw=boxedge,fill=white,line width=0.5pt,minimum height=6.5mm,inner sep=4pt,align=center},
  pill/.style={rounded corners=3pt,minimum height=6mm,inner sep=3pt,align=center,font=\sffamily\footnotesize},
  ansgood/.style={circle,draw=good,line width=0.9pt,fill=good!12,minimum size=7.2mm,inner sep=0pt,text=good,font=\sffamily\bfseries\scriptsize},
  ansbad/.style={circle,draw=warm,line width=0.9pt,fill=warm!12,minimum size=7.2mm,inner sep=0pt,text=warm,font=\sffamily\bfseries\scriptsize},
  lbl/.style={font=\sffamily\scriptsize\itshape,text=black!65},
  flow/.style={->,draw=black!55,line width=0.7pt},
  stem/.style={-,draw=black!55,line width=0.7pt},
  qtext/.style={align=left,font=\sffamily\scriptsize,text=black!80}
]

\node[imgbox] (img) at (0,0.15)
  {\includegraphics[width=22mm]{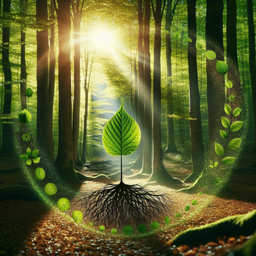}};
\node[qtext,right=2.5mm of img.east,anchor=west,text width=42mm] (q)
  {\textbf{ViLP-F}\\[0.4mm]
   \textit{A seed grows into a tree.}\\
   What grows into a tree in the image?};
\node[card] (sample) at (7.35,0.15) {Sample $N{=}8$\\reasoning traces};
\draw[flow] (q.east) -- (sample.west);

\node[ansbad] (c1) at (9.55,0.72) {seed};
\node[ansbad] (c2) at (10.40,0.72) {seed};
\node[ansbad] (c3) at (11.25,0.72) {seed};
\node[ansgood](c4) at (9.55,-0.30){leaf};
\node[ansgood](c5) at (10.40,-0.30){leaf};
\draw[flow] (sample.east) -- (9.05,0.20);
\node[lbl,align=left,anchor=west] at (11.80,0.72) {prior-driven\\[-0.15em]{\normalfont\fontsize{5}{6}\selectfont seed$\times$5}};
\node[lbl,align=left,anchor=west] at (11.80,-0.30){grounded\\[-0.15em]{\normalfont\fontsize{5}{6}\selectfont leaf$\times$3}};

\draw[black!15,line width=0.5pt] (-1.2,-1.25) -- (13.6,-1.25);

\node[pill,fill=black!6,draw=black!25,text width=26mm] (mv) at (1.1,-2.10)
  {\textbf{Majority vote}\\\scriptsize never re-reads $x$};
\node[ansbad,minimum size=8.5mm,font=\sffamily\bfseries\footnotesize] (mvpick) at (1.1,-3.65) {seed};
\node[lbl,text=warm,below=0.8mm of mvpick,align=center] {picks frequent\\$\Rightarrow$ \textbf{wrong}};
\draw[flow] (mv.south) -- (mvpick.north);

\node[pill,fill=good!10,draw=good,text=good,minimum width=52mm] (pgs) at (7.6,-2.00)
  {\textbf{\method{} (ours)}: re-score under edits of $x$};

\node[imgbox] (p1) at (5.5,-3.55)
  {\includegraphics[width=11mm]{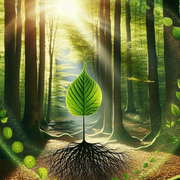}};
\node[imgbox] (p2) at (7.6,-3.55)
  {\includegraphics[width=11mm]{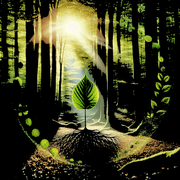}};
\node[imgbox] (p3) at (9.7,-3.55)
  {\includegraphics[width=11mm]{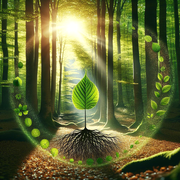}};
\node[lbl,below=0.8mm of p1] {crop};
\node[lbl,below=0.8mm of p2] {mask bg};
\node[lbl,below=0.8mm of p3] {jitter};

\coordinate (phub) at ([yshift=5mm]p2.north);
\draw[stem] (pgs.south) -- (phub);
\draw[flow] (phub) -| (p1.north);
\draw[flow] (phub) -- (p2.north);
\draw[flow] (phub) -| (p3.north);

\node[ansgood,minimum size=6.8mm] (v1) at (5.5,-5.05) {leaf};
\node[ansgood,minimum size=6.8mm] (v2) at (7.6,-5.05) {leaf};
\node[ansgood,minimum size=6.8mm] (v3) at (9.7,-5.05) {leaf};

\node[ansgood,minimum size=9.5mm,font=\sffamily\bfseries\footnotesize] (pgspick) at (7.6,-6.40) {leaf};
\draw[flow,draw=good!70] (v1.south) -- (pgspick.north west);
\draw[flow,draw=good!70] (v2.south) -- (pgspick.north);
\draw[flow,draw=good!70] (v3.south) -- (pgspick.north east);
\node[lbl,text=good,below=0.8mm of pgspick,align=center]
  {``leaf'' stable across views $\Rightarrow$ \textbf{correct}};

\end{tikzpicture}}
\captionof{figure}{PGS on Real ViLP-F example (Qwen2.5-VL-7B). A VLM samples $N{=}8$ traces for an
image--question pair whose prompt plants a language prior
(\textit{``A seed grows into a tree''}) while the image-grounded answer is
``leaf''. The prior guess ``seed'' is most
\emph{frequent} ($5$ vs.\ $3$), so \textbf{majority voting picks it and is wrong}---it
never re-reads the image. \method{} (ours) instead re-scores each candidate by adding
re-rendering $x$ under label-preserving edits (center crop, background mask,
photometric jitter): ``leaf'' stays modal across views.
So \method{} might selects the stable, image-grounded answer. Majority voting is the
empty-perturbation special case of \method{} (Section~\ref{sec:method}).}
\label{fig:teaser}
\end{center}

\section{Introduction}
\label{sec:intro}

Sampling many candidate solutions at inference and selecting a good one (test-time scaling) has become one of the most reliable ways to improve reasoning in large
language models (LLMs)~\citep{wei2022cot,wang2023selfconsistency,snell2024scaling,brown2024bon}.
The recipe assumes a usable \emph{selection signal}: either many samples agree on the
right answer (self-consistency / majority voting) or the model can verify its own
candidates well enough to pick the best one (best-of-$N$ with self-verification).

This recipe transfers poorly to vision-language models, and recent work converges on one
cause. For RL-tuned VLMs on visual math, majority voting beats verification-centric
selection and the self-correction ``aha moment'' yields no reliable
gain~\citep{ahamoment2026}. VLMs often attend to the correct evidence yet still answer
wrongly (``seeing but not believing''~\citep{seeingbelieving2025}) and can hold a fixed
answer while their internal representation drifts under label-preserving
edits~\citep{sameanswer2026}, so grounded and ungrounded answers are indistinguishable at
the output level and output stability is not evidence of grounding. Self-consistency
assumes correctness correlates with answer frequency, which fails when spurious paths
dominate~\citep{ttssurvey2025}; and across seven VLMs, single-model majority voting yields
only modest, chain-of-thought-dependent gains that vanish once outputs are
correlated~\citep{diversitymatters2026}. The common thread: the selection layer cannot tell
an image-grounded answer from a confident guess, because the usual signals (frequency,
verbalized confidence) never test whether the answer depends on the pixels. The selection
layer, not the generation layer, is where vision-language test-time scaling leaks.

The field's responses have largely gone in two directions, both of which leave the
core problem untouched for a practitioner with modest resources. One trains the
deficiency away with reinforcement learning or tool-use curricula so the model learns
when to re-examine visual
evidence~\citep{vtoolr1_2026,astra2026,valor2026}; this needs training compute and
data many groups do not have. The other intervenes inside a \emph{single} generation,
for example by masking deep-layer attention toward evidence
regions~\citep{seeingbelieving2025} or adding an external verifier or process reward
model that scores candidates~\citep{valor2026}; these need access to model internals,
a second trained network, or supervision. Neither gives a label-free, training-free
fix at the \emph{selection} layer that runs on a single consumer GPU, which is
precisely the regime in which test-time scaling is most attractive.

We turn this diagnosis into a design constraint: \emph{if the usual selection signals
fail because they never test whether an answer depends on the image, then a corrective
signal must be one that cannot be computed without the image.} A natural candidate,
which we call \methodlong{} (\method{}), scores each of the $N$ sampled answers by
re-rendering the visual input under a set of \emph{label-preserving
perturbations}, transformations that should not change the correct answer to a
genuinely image-grounded question (cropping toward question-relevant regions, masking
plausibly irrelevant background, mild photometric or geometric jitter), and measuring
how stable the model's support for that answer is across these views. The intuition is
that an answer re-derived from many perturbed views is grounded in the pixels, whereas
an answer that survives only on the original view is the fingerprint of a language-prior
guess. Majority voting is the empty-perturbation special case of \method{}, so this is a
strict generalization of the standard rule rather than a competing heuristic
(Section~\ref{sec:method}).

\method{} inherits none of the costs that make trained fixes inaccessible (no training,
labels, reward model, or second network, at a constant multiple of ordinary best-of-$N$
cost), so if a purely inference-time grounding signal can improve selection, \method{} is
where it should show. We therefore ask a sharper question than ``does \method{} beat majority
voting?'': \emph{does the perturbation signal add anything once the comparison is fair?}

The comparison is not fair by default, and this is the crux. \method{}'s perturbation-side
draws are \emph{short, answer-only} samples with no chain of thought, whereas plain majority
voting aggregates only the $N$ long CoT samples, so any gain of \method{} over MV conflates the
perturbation reweighting we care about with a switch in decoding format (CoT${\to}$short) that
we do not, and that format is not incidental for visual
tasks~\citep{ttsvlm2026,efftts_vlm2025}. We isolate the two with a \emph{format-matched control}
(MatchedCtrl): reuse the $N$ CoT answers and add the same number of short, no-CoT answers, but drawn from
the \emph{original} image with no perturbation. MatchedCtrl spends \method{}'s exact extra budget in its
exact decoding style and changes only whether those short answers pass through perturbed views,
so comparing \method{} to MatchedCtrl rather than to CoT-only MV tests grounding rather than format.

Our measurements across four automatically scored benchmarks---with Qwen as the
headline model and LLaVA-OneVision in the selector tables---return a
negative result. Against plain MV, \method{} appears to help substantially (up to $+31.8$
points on TextVQA; three-seed mean), but against MatchedCtrl it shows no reliable advantage on any
benchmark, including the vision-required ViLP, and no category yields a significant gain.
The preserve/destroy stability gap the method is built on is real and image-dependent (up to
$+0.48$), yet its per-instance value does not predict when \method{} wins. Perturbation consistency is thus at best a partial
diagnostic of visual dependence, not a usable label-free selection signal once decoding format
is held fixed.

\medskip\noindent\textbf{Contributions.}
\textbf{(1)~A decoding-format confound and its control:} comparing a perturbation- or
consistency-based selection rule against chain-of-thought-only majority voting conflates the
selection mechanism with a decoding-format change, and our format-matched control (MatchedCtrl), the
same short, no-CoT budget spent on the unperturbed image, isolates it, naming a cause distinct
from the diversity~\citep{diversitymatters2026} and internal-structure~\citep{arbiter2026}
accounts of weak VLM selection.
\textbf{(2)~A negative result under that control:} across four benchmarks (Qwen headline;
LLaVA-OneVision in matched-budget selectors), \method{}, a label-free, training-free rule that
strictly generalizes majority voting, gives no advantage over MatchedCtrl on any Qwen headline
benchmark, including the vision-required ViLP, and a real, image-dependent stability gap
(up to $+0.48$) does not predict when it wins.
\textbf{(3)~An experiment-facing isolation protocol:} we formalize the information and
budget difference among MV, \method{}, and MatchedCtrl, and state what would count as evidence that
perturbation reweighting supplies visual grounding at the selection layer
(Section~\ref{sec:isolation}).
\textbf{(4)~Released measurement and audit tooling} (the \method{} score, stability gap, paired
bootstrap CIs), so the format-matched comparison can be reused to
audit perturbation- and consistency-based selection claims for VLM test-time scaling.

\section{Related Work}
\label{sec:related}

\paragraph{Test-time scaling and self-verification.}
LLM test-time scaling selects among many samples via majority vote or best-of-$N$
verification~\citep{wang2023selfconsistency,brown2024bon,snell2024scaling}. The same
recipe is attractive for VLMs, yet the selection layer is fragile: for RL-tuned visual
math models, majority voting beats verification-centric selection and self-correction
gains are unreliable~\citep{ahamoment2026}, while self-refinement often
\emph{degrades} open VLMs~\citep{limitsgains2025}. Adaptive compute allocation and
trained multi-view policies~\citep{avis2026,sparc2026,mindjourney2025,astra2026}
improve \emph{generation}---when to look again, which crop to take---rather than how to
aggregate already-drawn candidates. \method{} is orthogonal: it leaves the generator
unchanged and only reweights answers at selection time, with no training and no second
network.

\paragraph{Perturbation signals and confounds.}
Input perturbations already appear as decoding interventions (e.g., contrastive
decoding~\citep{leng2024vcd,crg2024}) and as confidence or consistency
signals~\citep{zoomconsistency2026,ttcons2026}. Closest neighbors differ in what they
require (attention access, trained modules) or in what they claim
(generation-time grounding vs.\ selection-time reweighting); a compact design-properties
comparison is in the Supplementary Material~\citep{ascmqra2026,seeingbelieving2025}.
Parallel accounts of weak VLM selection emphasize sample
diversity~\citep{diversitymatters2026}, internal commitment versus
correctness~\citep{arbiter2026}, and the decoding
substrate~\citep{mlingtts2026,ttsvlm2026,efftts_vlm2025} (short answers often dominate
verbose CoT on visual tasks). We add a \emph{format-isolation} account: when a
perturbation-based selector spends its extra budget as short, no-CoT draws, a gain over
CoT-only majority voting can be almost entirely a CoT${\to}$short effect that a
format-matched control (MatchedCtrl) removes. Under that control, preserve/destroy stability
remains useful as an uncertainty \emph{diagnostic}~\citep{vluncertainty2024}, but not as
a selector.

\section{Method}
\label{sec:method}

\subsection{Setup and notation}
A VLM defines a distribution $p_\theta(a \mid x, q)$ over answers $a$ given an image
$x \in \mathcal{X}$ and a question $q$. Test-time scaling draws $N$ candidates
$a_1,\dots,a_N \sim p_\theta(\cdot \mid x, q)$ (with chain-of-thought, then extracting
the final answer) and applies a selection function $S(\{a_i\}, x, q) \to \hat a$.
Let $\mathrm{ans}(\cdot)$ map a sampled generation to its normalized final answer and
let $\mathcal{A} = \{\mathrm{ans}(a_i)\}$ be the set of distinct candidate answers.
Throughout, the image $x$ and question $q$ are fixed for an instance; what changes across
selectors is which additional draws are taken and how they enter $S$.

\paragraph{Majority voting.} The standard rule is
$S_{\mathrm{MV}} = \arg\max_{c \in \mathcal{A}} \sum_{i} \mathbf{1}[\mathrm{ans}(a_i)=c]$.
Each sample $a_i$ is generated conditioned on the image $x$, so MV \emph{does} use vision
at generation time; the aggregation step itself only counts extracted answers and does
not re-read $x$. That is a limitation of the \emph{selection} layer, not a claim that MV
ignores the image. Separately, \citet{ahamoment2026} show that self-verification-based
selection fails to integrate visual evidence effectively and is outperformed by MV---a
generation--verification gap that motivates seeking selection signals beyond unverified
self-checks, while still leaving open whether a perturbation-based reweighting can beat a
format-matched control.

\subsection{Label-preserving perturbations}
\begin{definition}[Label-preserving perturbation set]
A perturbation set $\mathcal{T} = \{t_1,\dots,t_M\}$ is a finite collection of maps
$t_m : \mathcal{X} \to \mathcal{X}$ such that for the true answer $a^\star$ to $(x,q)$,
$a^\star$ remains the correct answer to $(t_m(x), q)$ for all $m$. Examples used here:
(i) crops toward question-relevant regions proposed by a
cheap saliency heuristic; (ii) masking of background regions unlikely to contain the
answer; (iii) mild photometric jitter (brightness/contrast) and small-angle rotation
or rescaling.
\end{definition}

Perturbations are \emph{label-preserving by construction}, not by assumption about the
model: a crop that still contains the evidence, a mask over background, or a brightness
change does not alter the ground-truth answer to a well-posed visual question. We make
this concrete and auditable in Section~\ref{sec:experiments} via the
label-\emph{destroying} control, which deliberately removes the evidence region and
must reduce \method{}'s support for the correct answer if the signal is genuinely
grounded.

\subsection{The selection rule}
For candidate answer $c \in \mathcal{A}$, define its \emph{grounded support}
\begin{equation}
\begin{aligned}
g(c) \;=\; &\underbrace{\textstyle\sum_{i} \mathbf{1}[\mathrm{ans}(a_i)=c]}_{\text{original-view votes}}\\[2pt]
&\;+\; \lambda \textstyle\sum_{m=1}^{M} w_m \, \rho\big(c \mid t_m(x), q\big),
\end{aligned}
\label{eq:support}
\end{equation}
where $\rho(c \mid t_m(x), q)$ is the model's \emph{re-derivation strength} for answer
$c$ under perturbed view $t_m(x)$, $w_m \ge 0$ weights perturbation $m$, and
$\lambda \ge 0$ trades off original-view agreement against perturbation stability.
We instantiate $\rho$ as the consistency score
\begin{equation}
\rho(c \mid t_m(x), q) = \tfrac{1}{K}\textstyle\sum_{k=1}^{K}
\mathbf{1}\big[\mathrm{ans}(a^{(m,k)})=c\big],
\end{equation}
with $a^{(m,k)} \sim p_\theta(\cdot \mid t_m(x), q)$:
we draw $K$ short samples per perturbed view and count how often they re-derive
$c$. The selected answer is
$\hat a = \arg\max_{c \in \mathcal{A}} g(c)$.

\begin{algorithm}[t]
\caption{\methodlong{} (\method{})}
\label{alg:pgs}
\begin{algorithmic}[1]
\Require image $x$, question $q$, model $p_\theta$, perturbations $\mathcal{T}$,
counts $N,K$, weight $\lambda$
\State sample $a_1,\dots,a_N \sim p_\theta(\cdot\mid x,q)$; \;
$\mathcal{A}\gets\{\mathrm{ans}(a_i)\}$
\For{each perturbation $t_m \in \mathcal{T}$}
  \State sample $a^{(m,1)},\dots,a^{(m,K)} \sim p_\theta(\cdot\mid t_m(x),q)$
  \State record $\rho(c\mid t_m(x),q)$ for every $c\in\mathcal{A}$
\EndFor
\State compute $g(c)$ by Eq.~\eqref{eq:support} for every $c\in\mathcal{A}$
\State \Return $\hat a = \arg\max_{c} g(c)$
\end{algorithmic}
\end{algorithm}

\begin{proposition}[Majority voting is a special case]
\label{prop:mv}
If $\mathcal{T} = \emptyset$ (equivalently $\lambda = 0$), then $g(c)$ reduces to the
original-view vote count and $\hat a = S_{\mathrm{MV}}$.
\end{proposition}

\noindent This is immediate from Eq.~\eqref{eq:support}. \method{} is therefore a
strict generalization of majority voting rather than a competing heuristic: the
original-view vote count is always present in $g(c)$, and the perturbation term only
re-weights candidates. We do not claim \method{} can never underperform majority
voting (a large $\lambda$ can in principle override a correct original-view
majority), so the choice of $\lambda$ matters.
Unless noted otherwise, all primary results use a fixed operating
point $\lambda{=}2$ (chosen once for the protocol, not computed per example). Separately,
we also evaluate a \emph{label-free} selection rule that does not look at ground truth:
split the $K$ samples of each perturbed view into two halves, run selection on each half,
and among the $\lambda$ values whose two halves agree on the chosen answer at least
$80\%$ of the time, take the \emph{largest}. Choosing the smallest such $\lambda$ would
trivially return $\lambda{=}0$, since majority voting always agrees with itself. That rule
is reported in SensAblations alongside the full $\lambda$ sweep and a label-aware oracle; it is
\emph{not} the $\lambda$ used to produce Table~\ref{tab:main-results}.

\subsection{Cost}
The full procedure is Algorithm~1. The generation cost is
$N{+}MK$ forward samples versus $N$ for plain best-of-$N$ / CoT-only majority voting.
Throughout the experiments we hold the total number of generations fixed at a matched
budget (default $N{+}MK{=}32$; Section~\ref{sec:experiments}), so differences among
\method{}, and MatchedCtrl are about \emph{how} those draws are spent rather than about spending
more. Perturbed-view samples are short (answer-only), so the $MK$ term stays cheap relative
to $N$ long CoT traces; MatchedCtrl spends that same short-answer mass on the original image.

\section{What Would Count as a Grounding Gain?}
\label{sec:isolation}
\label{sec:theory}

The empirical claim is not that \method{} can never help relative to \emph{some}
baseline, but that a widely used comparison is misleading, and that under the
comparison that isolates the intended mechanism the gain disappears. This section
fixes that comparison before the numbers appear.

\subsection{Three selectors, three information budgets}
Fix an instance $(x,q)$ and a generation budget $B{=}N{+}MK$. Write
$\mathrm{CoT}_N(x)$ for $N$ chain-of-thought samples on the original image and
$\mathrm{Short}_K(y)$ for $K$ short, no-CoT samples on an image $y$.
\begin{itemize}
  \item \textbf{MV} uses only $\mathrm{CoT}_N(x)$ and returns a majority vote.
  Cost $N$.
  \item \textbf{\method{}} uses $\mathrm{CoT}_N(x)$ together with
  $\{\mathrm{Short}_K(t_m(x))\}_{m=1}^{M}$ and returns $\arg\max_c g(c)$
  (Eq.~\eqref{eq:support}). Cost $N{+}MK$.
  \item \textbf{MatchedCtrl} uses $\mathrm{CoT}_N(x)$ together with
  $\mathrm{Short}_{MK}(x)$---the \emph{same} short-answer mass, still on the original
  image---and returns a majority vote on the pooled answers. Cost $N{+}MK$.
\end{itemize}
MV and \method{} differ in \emph{two} ways at once: decoding format of the extra draws
(CoT vs.\ short) and whether those draws see perturbed pixels. MatchedCtrl matches \method{} on
budget and short-answer format and differs \emph{only} in whether the short draws are
routed through $\mathcal{T}$. Therefore:
\begin{quote}
\emph{A gain of \method{} over MV is not evidence of perturbation grounding.}\\
\emph{A gain of \method{} over MatchedCtrl would be.}
\end{quote}
Conversely, if $\mathrm{acc}(\method{})\approx\mathrm{acc}(\mathrm{MatchedCtrl})$, the
perturbation term is not buying selection accuracy beyond spending the same
short-answer budget on the original view. That is the decision-relevant null for
contribution~(2).

\subsection{Diagnostic vs.\ routing; operating point}
Even under a null against MatchedCtrl, the perturbation channel may still track visual content.
We separate two roles:
\begin{itemize}
  \item \textbf{Diagnostic (StabilityGap/BlankAblation).} The preserve/destroy stability gap, and the collapse
  of \method{} when perturbation inputs are blanked, test whether $\rho$ depends on the
  image. A large gap can coexist with a selection null.
  \item \textbf{Routing (RoutingTest).} For the same instances, does a larger gap predict a larger
  per-example gain $\mathbf{1}[\method{}]-\mathbf{1}[\mathrm{MatchedCtrl}]$? If not, the gap is not
  a usable switch for when to trust perturbation weighting over MatchedCtrl.
\end{itemize}
A \emph{grounding gain at the selection layer} would require (i)~a reliably positive
$\Delta(\method{}{-}\mathrm{MatchedCtrl})$ overall or in a pre-specified slice, and/or (ii)~a
monotone routing relationship in RoutingTest. Section~\ref{sec:experiments} tests both and finds
neither.
Primary tables fix $\lambda{=}2$, pool \texttt{union}, and vote weight $1$, holding total
generations at $N{+}MK{=}32$ unless noted. Sensitivity to $\lambda$, $M$, $K$, and
perturbation family is reported in SensAblations; those ablations do not overturn the MatchedCtrl null.

\section{Experiments}
\label{sec:experiments}

We evaluate the isolation protocol of Section~\ref{sec:isolation}: whether
\method{} improves over the format-matched control MatchedCtrl, and whether the stability gap
routes that comparison---not merely whether \method{} beats CoT-only majority voting.
The design therefore reports MV (for the familiar but confounded contrast), MatchedCtrl (for the
decision-relevant null), and the 	
StabilityGap/BlankAblation/RoutingTest diagnostics that separate ``the signal sees the
image'' from ``the signal improves selection.''
We evaluate on four benchmarks that vary in how much the answer depends on the image:
TextVQA~\citep{textvqa2019}, MATH-Vision~\citep{mathvision2024}, MMMU~\citep{mmmu2024}, and
ViLP~\citep{vilp2025}, the last of which pairs a language-prior-aligned answer with a
vision-required answer for each question and is the natural stress test for a
grounding-at-selection claim.
Coverage uses \textbf{two} open VLMs---\textbf{Qwen2.5-VL-7B-Instruct}~\citep{qwen25vl2025}
and \textbf{LLaVA-OneVision-7B}~\citep{llavali2024}.
The \emph{headline} mechanism table (Table~\ref{tab:main-results}) and the StabilityGap/	
BlankAblation/	
SensAblations/RoutingTest
analyses report Qwen three-seed means for a single readable story; matched-budget
selector comparisons and reproducibility tables report \emph{both} models
(Table~\ref{tab:r2-budget-32}, Table~\ref{tab:r2-budget-16}; and compute tables in the Supplementary Material).
Unless stated otherwise, \method{} is recomputed offline with candidate pool
\texttt{union}, vote weight $\mathrm{vw}{=}1$, and $\lambda{=}2$ on saved runs.
Accuracy is hard-match rate in percent (TextVQA: any-annotator match).

\begin{table}[t]
\centering
\small
\setlength{\tabcolsep}{2.6pt}
\caption{\textbf{Main results on Qwen2.5-VL-7B-Instruct}. Mean hard-match accuracy over seeds $\{0,23,42\}$. \method{} appears to beat
plain majority voting (MV) by up to $+31.8$\,pp (column $\Delta$), yet the format-matched
control MatchedCtrl (shaded)---the same short, no-CoT budget on the \emph{original} image---tracks
\method{} within seed noise on every benchmark (TextVQA: MatchedCtrl still ahead). Accuracy is
hard-match rate (\%; TextVQA any-annotator match); \method{} uses \texttt{union}, vote weight
$1$, $\lambda{=}2$.}
\label{tab:main-results}
\renewcommand{\arraystretch}{1.18}
\begin{threeparttable}
\begin{tabular}{@{}l
  S[table-format=3.0]
  S[table-format=2.1]
  S[table-format=2.1]
  >{\columncolor{hlcol}}S[table-format=2.1]
  S[print-implicit-plus,table-format=+2.1]
  S[print-implicit-plus,table-format=+1.3]@{}}
\toprule
 & & \multicolumn{3}{c}{\bfseries Accuracy (\%)} & & \\
\cmidrule(lr){3-5}
\multicolumn{1}{@{}l}{\bfseries Bench.} & \headrowc{$N$} & \headrowc{MV} &
\headrowc{\method{}} & \multicolumn{1}{c}{\cellcolor{hlcol}\bfseries MatchedCtrl\tnote{a}} &
\headrowc{$\Delta$\tnote{b}} & \headrowc{Gap\tnote{c}} \\
\midrule
TextVQA & 300 & 54.4 & 86.2 & 87.6 & +31.8 & +0.478 \\
MATH-V  & 299 & 25.6 & 26.0 & 25.3 &  +0.4 & +0.025 \\
MMMU    & 900 & 49.5 & 50.3 & 50.1 &  +0.7 & +0.071 \\
ViLP    & 600 & 53.2 & 52.3 & 53.7 &  -0.9 & +0.445 \\
\bottomrule
\end{tabular}
\begin{tablenotes}[flushleft]\scriptsize
\item[a] \textbf{MatchedCtrl}, format-matched control: reuse the $N$ CoT answers $+$ $M{\cdot}K$ short
no-CoT answers on the original image ($N{+}MK{=}32$).
\item[b] $\Delta=\mathrm{acc}(\method{})-\mathrm{acc}(\mathrm{MV})$, from seed-mean
accuracies.
\item[c] Mean Gap $\rho_{\text{preserve}}-\rho_{\text{destroy}}$ over seeds; ViLP gaps are
computed on non-prior (grounded) slots in the dump.
\end{tablenotes}
\end{threeparttable}
\end{table}

\subsection{Setup and controls}
\label{sec:exp-setup}

\paragraph{Protocols.}
For each example we draw $N$ original-view answers with chain-of-thought (CoT)
and, for \method{}, $K$ \emph{short, no-CoT} answers on each of $M$
label-preserving perturbations.
\method{} scores candidates by votes plus $\lambda$ times re-derivation support
$\rho$ estimated from the perturbation views.
We report:
\begin{itemize}
  \item \textbf{MV:} majority vote over the $N$ original-view \emph{CoT} samples only.
  \item \textbf{\method{}:} $\rho$-weighted selection as above.
  \item \textbf{MVLift:} paired difference $\method{}-\mathrm{MV}$ (percentage points).
  \item \textbf{MatchedCtrl:} format-matched control that reuses the $N$ CoT answers and adds
        $M{\cdot}K$ short no-CoT answers on the \emph{original} image ($N{+}MK{=}32$),
        matching \method{}'s extra budget and decoding style with no perturbation.
  \item \textbf{StabilityGap:} mean paired stability gap
        $\rho_{\mathrm{preserve}}-\rho_{\mathrm{destroy}}$ on the MV-selected
        candidate under label-destroying crops.
\end{itemize}
Comparing \method{} to MV confounds perturbation weighting with the CoT${\leftrightarrow}$short
format change; comparing \method{} to MatchedCtrl isolates whether routing those short answers through
perturbed views and $\rho$ helps beyond keeping them on the original image.

\paragraph{Baselines.}
Our comparison isolates the perturbation mechanism against the format-matched
control (MatchedCtrl), the decisive test for our claim; it is not a survey of selectors.
Still, under the same matched-budget protocol on each VLM we evaluate standard
label-free alternatives---self-certainty (SC), SC with Borda
voting~\citep{selfcertainty2025}, confidence-weighted
self-consistency~\citep{cisc2025}, and mean token-entropy selection---against
MV, MatchedCtrl, and \method{} at $N{+}MK\in\{16,32\}$
(Table~\ref{tab:r2-budget-32}, Table~\ref{tab:r2-budget-16}).
Both LLaVA-OneVision-7B and Qwen appear in these selector tables, so the MatchedCtrl null
is not an artifact of a single checkpoint family.
Confidence selectors use the matched $N{+}MK$ decode pool after teacher-force
rescoring; budget~$16$ is an offline subsample ($N{=}4$, $K{=}2$) of the
budget-$32$ dumps.
Consistent with the intro diagnosis, these selectors do not systematically beat
MatchedCtrl: SC and entropy often fall below MatchedCtrl, while SC+Borda and CISC typically track
MatchedCtrl within noise.
Where \method{} also fails to separate from MatchedCtrl, the null remains diversity /
format rather than grounding; where absolute numbers differ across models, the
gap is still relative to a control that already absorbs the strongest
confidence-based aggregators.

\begin{table}[t]
\centering
\caption{Label-free selectors vs.\ MV/MatchedCtrl/\method{} at matched budget
$N{+}MK{=}32$ (rescored confidence metrics; mean$\pm$std over seeds).}
\label{tab:r2-budget-32}
\setlength{\tabcolsep}{3.2pt}
\resizebox{\textwidth}{!}{%
\begin{tabular}{@{}llccccccc@{}}
\toprule
Benchmark & Model & MV & MatchedCtrl & SC & Entropy & SC+Borda & CISC & PGS \\
\midrule
MATH-Vision & LLaVA-OV-7B & 18.8$\pm$1.1 & 20.7$\pm$0.9 & 15.8$\pm$1.1 & 17.8$\pm$0.7 & 19.2$\pm$2.3 & 19.5$\pm$0.8 & 20.3$\pm$0.2 \\
MATH-Vision & Qwen2.5-VL-7B & 25.6$\pm$1.6 & 25.3$\pm$1.0 & 17.9$\pm$0.5 & 15.4$\pm$0.6 & 25.1$\pm$1.0 & 25.4$\pm$1.0 & 26.0$\pm$0.7 \\
MMMU & LLaVA-OV-7B & 47.4$\pm$0.5 & 47.6$\pm$0.6 & 43.9$\pm$0.3 & 45.7$\pm$0.3 & 47.5$\pm$0.7 & 47.9$\pm$0.7 & 48.3$\pm$0.7 \\
MMMU & Qwen2.5-VL-7B & 49.5$\pm$0.3 & 50.1$\pm$0.4 & 44.7$\pm$0.5 & 44.7$\pm$0.7 & 50.1$\pm$0.5 & 50.4$\pm$0.4 & 50.3$\pm$0.7 \\
TextVQA & LLaVA-OV-7B & 78.8$\pm$0.2 & 81.3$\pm$0.3 & 78.8$\pm$0.8 & 80.4$\pm$0.8 & 82.2$\pm$0.5 & 81.6$\pm$0.5 & 81.7$\pm$0.3 \\
TextVQA & Qwen2.5-VL-7B & 54.4$\pm$2.0 & 87.6$\pm$0.4 & 47.1$\pm$2.1 & 84.0$\pm$4.1 & 88.1$\pm$0.5 & 88.0$\pm$0.3 & 86.2$\pm$0.7 \\
ViLP & LLaVA-OV-7B & 49.9$\pm$0.3 & 49.4$\pm$0.3 & 50.4$\pm$0.5 & 49.1$\pm$0.1 & 50.4$\pm$0.9 & 50.1$\pm$0.7 & 49.5$\pm$0.7 \\
ViLP & Qwen2.5-VL-7B & 53.2$\pm$0.7 & 53.7$\pm$0.4 & 50.9$\pm$0.6 & 48.6$\pm$0.6 & 54.0$\pm$0.2 & 54.0$\pm$0.4 & 52.3$\pm$0.7 \\
\bottomrule
\end{tabular}%
}
\end{table}

\begin{table}[t]
\centering
\caption{Label-free selectors vs.\ MV/MatchedCtrl/\method{} at matched budget
$N{+}MK{=}16$ (offline subsample from budget-$32$ dumps; mean$\pm$std over seeds).}
\label{tab:r2-budget-16}
\setlength{\tabcolsep}{3.2pt}
\resizebox{\textwidth}{!}{%
\begin{tabular}{@{}llccccccc@{}}
\toprule
Benchmark & Model & MV & MatchedCtrl & SC & Entropy & SC+Borda & CISC & PGS \\
\midrule
MATH-Vision & LLaVA-OV-7B & 20.9$\pm$2.2 & 20.7$\pm$1.2 & 17.4$\pm$1.4 & 19.0$\pm$0.7 & 20.4$\pm$0.6 & 20.3$\pm$1.2 & 22.9$\pm$0.8 \\
MATH-Vision & Qwen2.5-VL-7B & 23.3$\pm$1.6 & 26.8$\pm$2.6 & 23.0$\pm$1.9 & 21.6$\pm$2.2 & 26.9$\pm$2.3 & 26.3$\pm$2.6 & 26.8$\pm$1.1 \\
MMMU & LLaVA-OV-7B & 48.2$\pm$1.2 & 48.3$\pm$0.6 & 44.8$\pm$0.1 & 47.1$\pm$0.4 & 47.6$\pm$0.4 & 48.2$\pm$0.9 & 48.8$\pm$1.1 \\
MMMU & Qwen2.5-VL-7B & 49.3$\pm$0.8 & 51.0$\pm$0.5 & 45.4$\pm$0.6 & 45.7$\pm$0.6 & 51.3$\pm$0.1 & 51.2$\pm$0.1 & 50.6$\pm$0.3 \\
TextVQA & LLaVA-OV-7B & 75.7$\pm$0.3 & 81.8$\pm$0.2 & 79.3$\pm$0.7 & 81.3$\pm$0.3 & 81.8$\pm$0.4 & 82.0$\pm$0.7 & 81.6$\pm$0.2 \\
TextVQA & Qwen2.5-VL-7B & 57.0$\pm$2.0 & 87.9$\pm$0.4 & 49.9$\pm$2.9 & 84.7$\pm$2.7 & 88.4$\pm$0.7 & 88.4$\pm$0.2 & 88.7$\pm$0.7 \\
ViLP & LLaVA-OV-7B & 49.8$\pm$0.3 & 49.5$\pm$0.4 & 50.5$\pm$0.2 & 49.4$\pm$0.2 & 50.1$\pm$0.8 & 50.1$\pm$0.6 & 48.9$\pm$0.8 \\
ViLP & Qwen2.5-VL-7B & 52.9$\pm$0.8 & 53.3$\pm$0.3 & 51.8$\pm$1.2 & 49.7$\pm$0.3 & 53.7$\pm$0.4 & 53.6$\pm$0.2 & 51.2$\pm$0.7 \\
\bottomrule
\end{tabular}%
}
\end{table}


\paragraph{Reproducibility and benchmarks.}
We evaluate TextVQA, MATH-Vision, MMMU, and ViLP (Score / ViLP-P / ViLP-F splits as
noted per table; ViLP StabilityGap gaps use non-prior grounded slots in the dumps, $N{=}600$ per
seed).
Decode settings, exact $(N,M,K)$, compute footprint for \emph{both} VLMs, dataset
versions/splits/licenses, and checkpoint revisions appear in the Supplementary Material. 
Default generation budget is $N{=}8$, $M{=}6$, $K{=}4$ ($N{+}MK{=}32$) on a single
NVIDIA RTX~4090.

\subsection{Main results}
\label{sec:exp-main}

Table~\ref{tab:main-results} is the headline comparison (three-seed means on Qwen dumps).
Against plain MV, \method{} gains substantially on TextVQA ($+31.8$\,pp) and is near-flat on
MATH-V, MMMU, and ViLP. That comparison is unfair in exactly the sense of
Section~\ref{sec:isolation}: MV aggregates only the $N$ \emph{CoT} samples, whereas both
\method{} and MatchedCtrl add $M{\cdot}K$ \emph{short, no-CoT} answers. The all-CoT MV pool is
therefore the weaker aggregator at a smaller effective budget in short-answer mass:
accuracy rises at matched $N{+}MK$ even when the extra draws never leave the original image.
MatchedCtrl spends those same short answers on the \emph{original} image and, under this
format-matched control, tracks \method{} within seed noise on every benchmark (TextVQA
$87.6$ vs.\ $86.2$; MATH-V $25.3$ vs.\ $26.0$; MMMU $50.1$ vs.\ $50.3$; ViLP $53.7$ vs.\
$52.3$), with MatchedCtrl still ahead on TextVQA. The MV$\to$\method{} lift is therefore the effect
of adding short no-CoT mass, not evidence that re-derivation support $\rho$ supplies visual
grounding at the selection layer; isolating the perturbation piece leaves no reliable
advantage.

The StabilityGap reinforces this reading rather than rescuing it. Large mean
preserve/destroy gaps on TextVQA and ViLP ($+0.478$ / $+0.445$) coexist with no win over
MatchedCtrl, while smaller gaps on MATH-V and MMMU ($+0.025$ / $+0.071$) likewise fail to separate
\method{} from MatchedCtrl. Preserve/destroy asymmetry is real and image-dependent, but its
magnitude does not identify when perturbation-weighted selection beats format-matched
short-answer MV. In other words, the diagnostic channel can fire without a routing gain
over MatchedCtrl---the distinction Section~\ref{sec:isolation} insists on.

\subsection{Ablations: the signal is visual but does not route selection}
\label{sec:exp-ablation}
\begin{table}[t]
\centering
\caption{\textbf{BlankAblation} (earlier single-seed Qwen dual-PGS run; not
present in the multi-seed dumps used for Table~\ref{tab:main-results}). Holding the
original-view CoT pool fixed and blanking the perturbation inputs collapses accuracy where
the signal is visual (TextVQA, ViLP-F, MMMU) but not on symbolic MATH-V. Absolutes are
\emph{not} aligned with the multi-seed main table; only the within-row comparison is
meaningful.}
\label{tab:e4}
\renewcommand{\arraystretch}{1.2}
\begin{tabular}{@{}l S[table-format=3.0] S[table-format=2.1] S[table-format=2.1] S[table-format=2.1]@{}}
\toprule
\multicolumn{1}{@{}l}{\bfseries Benchmark} & \headrowc{$N$} & \headrowc{MV} &
\headrowc{\method{}} & \headrowc{BlankAblation} \\
\midrule
TextVQA & 302 & 55.0 & 87.7 &  7.9 \\
MATH-V  & 304 & 22.4 & 26.3 & 23.4 \\
MMMU    & 900 & 49.9 & 50.2 & 23.3 \\
ViLP-F  & 266 & 48.9 & 46.2 &  0.4 \\
\bottomrule
\end{tabular}
\end{table}

Blanking the perturbation inputs (BlankAblation, Table~\ref{tab:e4}) collapses \method{}'s accuracy where
the signal is strong (TextVQA $87.7\!\to\!7.9$, ViLP-F $46.2\!\to\!0.4$, MMMU
$50.2\!\to\!23.3$) while symbolic MATH-V is unaffected, confirming the score is genuinely image-dependent rather than a format artifact of short decoding alone. Yet
image-dependence and the stability gap decouple (MMMU drops sharply while its multi-seed
mean gap is only $+0.071$), so the gap is only a partial measure of visual dependence:
BlankAblation can fail even when StabilityGap looks mild. Ablations over perturbation family, $\lambda$, $M$,
and $K$ (SensAblations, Supplementary Material: perturbation-family / $\lambda$-sweep /
$\hat\lambda$ tables; recomputed offline from the same multi-seed Qwen dumps as
Table~\ref{tab:main-results}) show \method{} is active almost only on OCR-heavy TextVQA
(crop/geometry-sensitive) while other benchmarks are flat, consistent with a
format-sensitive rather than a universal grounding effect. Neither ablation overturns
MatchedCtrl: where the signal is visually live, matched short answers on the original image still
absorb the lift.

\begin{figure}[t]
\centering
\includegraphics[width=0.55\linewidth]{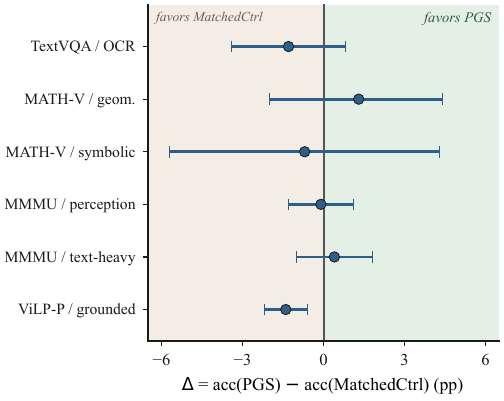}
\caption{\textbf{No category shows a significant gain of \method{} over the format-matched
control MatchedCtrl.} Per-category difference $\Delta=\text{acc}(\method{})-\text{acc}(\mathrm{MatchedCtrl})$
with 95\% bootstrap confidence intervals (Supplementary Material), pooled over seeds
$\{0,23,42\}$ on Qwen dumps. No interval lies entirely in the ``favors \method{}''
half-plane (shaded); the ViLP interval lies entirely below zero. Points are $\Delta$;
whiskers are the 95\% CI.}
\label{fig:e6-forest}
\end{figure}

\subsection{RoutingTest: where the gap predicts help (it does not)}
\label{sec:exp-e6}

We break results down by perception-heavy vs.\ text-heavy categories and correlate
per-instance gain $(\mathbf{1}[\method{}]-\mathbf{1}[\mathrm{MatchedCtrl}])$ with the StabilityGap; on ViLP we additionally report prior$\to$grounded flip rates relative to MatchedCtrl. These
analyses ask the routing question of Section~\ref{sec:isolation} directly: even if the
mean $\Delta(\method{}{-}\mathrm{MatchedCtrl})$ is near zero, a pre-specified slice or a monotone
gap$\to$gain relationship could still salvage a conditional use of $\rho$. The Supplementary Material category table and Figure~\ref{fig:e6-forest} close the isolation
protocol under the decision-relevant baseline: no category yields a significant positive
$\Delta(\method{}-\mathrm{MatchedCtrl})$ (every 95\% bootstrap CI overlaps zero). The stability gap
also fails to \emph{route} the selector: instance-level correlations between gap and gain
are near zero (Supplementary Material), gap quartiles are non-monotone, and \method{}'s
recoveries of MatchedCtrl errors on ViLP are rare (Supplementary Material). A large measured
re-derivation advantage therefore does not forecast when perturbation-weighted selection
beats format-matched short-answer MV. The gap remains a useful \emph{diagnostic} of
preserve/destroy asymmetry; it is not a reliable \emph{routing} signal for the selector
studied here.

\section{Discussion}
\label{sec:discussion}

The controls converted a plausible mechanism into a measured null. StabilityGap and BlankAblation show that the re-derivation score tracks
visual content when the pixels are removed or destroyed; once the same short draws sit on
the original image, the perturbation term buys nothing at selection, and
the per-instance gap does not route the selector (RoutingTest). Diagnostic image dependence
is therefore not a routing gain over a format-matched control
(Section~\ref{sec:isolation}). Gains against CoT-only majority voting can be
almost entirely a decoding-format effect---here as large as $+31.8$\,pp on TextVQA---and
should not be read as evidence that perturbation consistency supplies visual grounding.
ViLP sharpens the point: even where language priors and image answers are split, MatchedCtrl still meets or beats \method{}, so a large
preserve/destroy gap coexists with a selection null rather than rescuing it. This isolates a format piece that coexisting 2026 accounts do not foreground in the same
way: diversity-limited self-consistency~\citep{diversitymatters2026} and
commitment-versus-correctness~\citep{arbiter2026} explain other failure modes of VLM
selection, while MatchedCtrl shows that a CoT${\leftrightarrow}$short confound can inflate apparent
perturbation gains even when sample diversity and internal stability are held aside.
Holding the \emph{number} of draws fixed is not enough if the extra draws
change answer format; the control must also hold format fixed on the original view.
Progress beyond MatchedCtrl will likely need an information source that answer-level consistency
alone does not supply---denser visual probes, process-level rewards, or
cross-model disagreement. We do \emph{not} show that a trained verifier,
or multi-model ensemble beats MatchedCtrl; whether such a signal recovers the gap remains
open, and any such claim must still clear a format-matched control.

\paragraph{Practical takeaways.} Any selection rule that spends extra draws in a different
decoding format than the MV baseline needs a format-matched control; without one, a format effect is easily misread as a grounding gain. Short-answer
aggregation on the original image is a strong, cheap baseline under matched budget, and
in our grid it is the hurdle that confidence-style and perturbation-weighted selectors
alike must clear. Perturbation consistency remains worth computing as a partial
\emph{diagnostic} of visual dependence (StabilityGap/BlankAblation)---useful for auditing whether a score
still sees the image---but not as a drop-in \emph{selector} once format is controlled.
Reporting MV lifts without a MatchedCtrl-style arm should be treated as incomplete for grounding-at-selection claims.

\section*{Limitations}
\label{sec:limitations}
Our conclusion is bounded, and stating the bounds is part of reading it honestly.
The evidence covers two open VLMs (LLaVA-OneVision-7B and Qwen2.5-VL-7B) on four
automatically scored benchmarks. A stronger family or larger decode budget could in principle show a benefit ours
do not; we do not claim universality beyond the measured grid.
Main results are aggregated over three decoding seeds ($\{0,23,42\}$) as
mean$\pm$std (Supplementary Material); paired \method{}-vs-MatchedCtrl comparisons
carry instance-level bootstrap CIs (RoutingTest), the right uncertainty for
the claim that no category shows a significant gain over MatchedCtrl. Three seeds cannot rule out rarer decoding regimes.
The perturbation families remain hand-designed heuristics (crop, mask, mild photometric
and geometric jitter); a learned set could carry more signal, but
would still need a MatchedCtrl-style control before any gain is attributed to grounding, and our
SensAblations sweeps already show that family choice does not overturn the null under matched
format.
We do not include a stronger-signal reference point (trained verifier or multi-model
ensemble) that beats MatchedCtrl, so recoverability under training is not demonstrated.
And the negative finding is about \emph{answer selection}: the stability gap remains a
partial diagnostic of visual dependence---none of which we test here for abstention or routing.

\section*{Ethics Statement}
\label{sec:ethics}
\method{} is inference-only and uses public benchmarks and open models, raising no
annotation- or participant-related concerns. Favoring image-grounded answers may reduce
confidently-wrong outputs that ignore the pixels, but a stability signal can be satisfied
by inputs that are stable yet wrong, so \method{} is not a correctness or safety guarantee
and does not replace verification in safety-critical use; it also inherits the underlying
model's biases and failure modes. Deploying perturbation probes still incurs extra decode
cost and can amplify whatever stereotypes or dataset artifacts the base VLM already
exhibits under short-answer sampling.
Because our headline finding is negative under MatchedCtrl, the main ethical risk of overstating
\method{} as a grounding selector is mitigated by the release itself: we release the
analysis code, perturbation specifications, and evaluation tooling with paired bootstrap
CIs, so the null against format-matched short-answer aggregation is reproducible.
We encourage downstream work that claims a grounding gain at selection to
publish the same MatchedCtrl-style contrast rather than MV-only lifts. The authors used AI-based tools for assistance with language editing and 
software debugging during the development of this work. All AI-generated 
suggestions were reviewed, verified, and modified by the authors.

\begin{availability}
PGS implementation and audit tooling (paired bootstrap CIs) are released on \url{https://github.com/KurbanIntelligenceLab/PGS-Code-Audit-Tooling.git}.
\end{availability}

\begin{conflicts}
The authors declare no competing interests.
\end{conflicts}

\putbib
\end{bibunit}

\clearpage
\makeatletter
\renewcommand{\appendix}{%
  \KIL@origappendix
  \setcounter{section}{0}\setcounter{figure}{0}\setcounter{table}{0}\setcounter{equation}{0}%
  \renewcommand{\thesection}{\Alph{section}}%
  \renewcommand{\thefigure}{A\arabic{figure}}%
  \renewcommand{\thetable}{A\arabic{table}}%
  \renewcommand{\theequation}{A\arabic{equation}}}
\makeatother
\appendix
\begin{bibunit}[unsrtnat]
\raggedbottom
\setlength{\textfloatsep}{6pt plus 1pt minus 1pt}
\setlength{\floatsep}{6pt plus 1pt minus 1pt}
\setlength{\intextsep}{6pt plus 1pt minus 1pt}
\captionsetup{font=small,skip=4pt}
\renewcommand{\arraystretch}{1.08}
\section{Supplementary Material}
\label{app:tables}

\subsection{Reproducibility checklist tables}
\begin{center}
\footnotesize
\captionof{table}{Decode settings used in all reported runs.}
\label{tab:r3-decode}
\begin{tabular}{@{}lc@{}}
\toprule
Setting & Value \\
\midrule
cot\_temperature & 0.7 \\
short\_temperature & 0.7 \\
perturbed\_temperature & 0.7 \\
top\_p & 0.95 \\
cot\_max\_new\_tokens & 256 \\
short\_max\_new\_tokens & 64 \\
do\_sample & True \\
\bottomrule
\end{tabular}
\end{center}
\vspace{0.35em}

\begin{center}
\footnotesize
\captionof{table}{Generation budget $(N,M,K)$ (uniform across benchmarks in these dumps).}
\label{tab:r3-budget}
\setlength{\tabcolsep}{4pt}
\begin{tabular}{@{}lcccc@{}}
\toprule
$N$ & $M$ & $K$ & $N{+}MK$ & per-family $K$ \\
\midrule
8 & 6 & 4 & 32 &
\makecell[l]{center\_crop=4, saliency\_crop=4,\\
background\_mask=4, photometric=4,\\
rotation=4, rescale=4} \\
\bottomrule
\end{tabular}
\end{center}
\vspace{0.35em}
\begingroup
\scriptsize
\setlength{\tabcolsep}{3.5pt}
\begin{longtable}{@{}lllcccc@{}}
\caption{Compute footprint per run.
Each benchmark$\times$seed has one row per VLM.}
\label{tab:r3-compute}\\
\toprule
Benchmark & Model & Seed & GPU & Torch / Transformers & \#ex & GPU-h \\
\midrule
\endfirsthead
\multicolumn{7}{c}{\footnotesize\textit{Table~\thetable{} continued from previous page}}\\[0.3em]
\toprule
Benchmark & Model & Seed & GPU & Torch / Transformers & \#ex & GPU-h \\
\midrule
\endhead
\midrule
\multicolumn{7}{r}{\emph{Continued on next page}}\\
\endfoot
\bottomrule
\endlastfoot
MATH-Vision & LLaVA-OV-7B & 0 & RTX 4090 & 2.13.0+cu130 / 5.14.1 & 304 & 2.96 \\
MATH-Vision & Qwen2.5-VL-7B & 0 & RTX 4090 & 2.6.0+cu124 / 5.14.1 & 299 & 4.63 \\
MATH-Vision & LLaVA-OV-7B & 23 & RTX 4090 & 2.13.0+cu130 / 5.14.1 & 304 & 2.60 \\
MATH-Vision & Qwen2.5-VL-7B & 23 & RTX 4090 & 2.6.0+cu124 / 5.14.1 & 283 & 4.48 \\
MATH-Vision & LLaVA-OV-7B & 42 & RTX 4090 & 2.6.0+cu124 / 5.14.1 & 304 & 2.95 \\
MATH-Vision & Qwen2.5-VL-7B & 42 & RTX 4090 & 2.6.0+cu124 / 5.14.1 & 299 & 4.66 \\
MMMU & LLaVA-OV-7B & 0 & RTX 4090 & 2.13.0+cu130 / 5.14.1 & 900 & 5.60 \\
MMMU & Qwen2.5-VL-7B & 0 & RTX 4090 & 2.13.0+cu130 / 5.14.1 & 900 & 6.10 \\
MMMU & LLaVA-OV-7B & 23 & RTX 4090 & 2.6.0+cu124 / 5.14.1 & 900 & 5.79 \\
MMMU & Qwen2.5-VL-7B & 23 & RTX 4090 & 2.6.0+cu124 / 5.14.1 & 900 & 4.88 \\
MMMU & LLaVA-OV-7B & 42 & RTX 4090 & 2.13.0+cu130 / 5.14.1 & 900 & 5.95 \\
MMMU & Qwen2.5-VL-7B & 42 & RTX 4090 & 2.6.0+cu124 / 5.14.1 & 900 & 6.40 \\
TextVQA & LLaVA-OV-7B & 0 & RTX 4090 & 2.6.0+cu124 / 5.14.1 & 300 & 2.74 \\
TextVQA & Qwen2.5-VL-7B & 0 & RTX 4090 & 2.6.0+cu124 / 5.14.1 & 300 & 1.93 \\
TextVQA & LLaVA-OV-7B & 23 & RTX 4090 & 2.6.0+cu124 / 5.14.1 & 300 & 2.93 \\
TextVQA & Qwen2.5-VL-7B & 23 & RTX 4090 & 2.13.0+cu130 / 5.14.1 & 300 & 1.90 \\
TextVQA & LLaVA-OV-7B & 42 & RTX 4090 & 2.6.0+cu124 / 5.14.1 & 300 & 2.74 \\
TextVQA & Qwen2.5-VL-7B & 42 & RTX 4090 & 2.13.0+cu130 / 5.14.1 & 300 & 1.92 \\
ViLP & LLaVA-OV-7B & 0 & RTX 4090 & 2.6.0+cu124 / 5.14.1 & 600 & 8.07 \\
ViLP & Qwen2.5-VL-7B & 0 & RTX 4090 & 2.6.0+cu124 / 5.14.1 & 600 & 3.66 \\
ViLP & LLaVA-OV-7B & 23 & RTX 4090 & 2.13.0+cu130 / 5.14.1 & 600 & 7.88 \\
ViLP & Qwen2.5-VL-7B & 23 & RTX 4090 & 2.13.0+cu130 / 5.14.1 & 600 & 3.84 \\
ViLP & LLaVA-OV-7B & 42 & RTX 4090 & 2.6.0+cu124 / 5.14.1 & 600 & 7.62 \\
ViLP & Qwen2.5-VL-7B & 42 & RTX 4090 & 2.6.0+cu124 / 5.14.1 & 600 & 3.91 \\
\end{longtable}
\endgroup

\clearpage
\subsection{Additional results}

\begin{center}
\footnotesize
\captionof{table}{\textbf{Per-category difference of \method{} vs.\ the format-matched control MatchedCtrl,
with 95\% bootstrap confidence intervals.} $\Delta=\mathrm{acc}(\method{})-\mathrm{acc}(\mathrm{MatchedCtrl})$.
Instances pooled over seeds $\{0,23,42\}$ on Qwen dumps; $N$ is the mean count per seed.
No category's CI lies entirely above zero (no significant gain). Visualized in the main-paper forest plot (Figure~\ref{fig:e6-forest}). The gap column is the preserve$-$destroy stability gap.
MATH-V perception vs.\ text uses subject tags; MMMU uses \texttt{category\_bucket}
(perception-heavy / text-heavy; ``other'' omitted). $\Delta$ from unrounded accuracies.}
\label{tab:e6-cat-e2}
\renewcommand{\arraystretch}{1.08}
\begin{threeparttable}
\resizebox{\textwidth}{!}{%
\begin{tabular}{@{}l l
  S[table-format=3.0]
  S[table-format=2.1]
  S[table-format=2.1]
  S[print-implicit-plus,table-format=+2.1]
  c
  S[print-implicit-plus,table-format=+1.3]@{}}
\toprule
\multicolumn{1}{@{}l}{\bfseries Benchmark} & \bfseries Category & \headrowc{$N$} &
\headrowc{MatchedCtrl} & \headrowc{\method{}} & \headrowc{$\Delta$} &
\multicolumn{1}{c}{\bfseries 95\% CI} & \headrowc{Gap} \\
\midrule
TextVQA & perception / OCR   & 300 & 87.6 & 86.2 & -1.3 & {$[-3.4,\,+0.8]$}  & +0.478 \\
MATH-V  & perception / geom. & 205 & 21.5 & 22.8 & +1.3 & {$[-1.8,\,+4.4]$}  & +0.010 \\
MATH-V  & text / symbolic    &  94 & 33.7 & 33.0 & -0.7 & {$[-5.7,\,+4.3]$}  & +0.057 \\
MMMU    & perception-heavy   & 540 & 47.1 & 47.0 & -0.1 & {$[-1.2,\,+1.2]$}  & +0.070 \\
MMMU    & text-heavy         & 330 & 52.8 & 53.2 & +0.4 & {$[-1.0,\,+1.8]$}  & +0.067 \\
ViLP-P  & grounded (Score)   & 600 & 53.7 & 52.3 & -1.4 & {$[-2.2,\,-0.6]$}  & +0.445 \\
\bottomrule
\end{tabular}%
}
\begin{tablenotes}[flushleft]\footnotesize
\item MatchedCtrl uses $N{+}M{\cdot}K{=}32$ original-view samples (reuse $N$ CoT $+$ $M{\cdot}K$ short
no-CoT answers). Per-category accuracies are computed within the categorized subset.
\end{tablenotes}
\end{threeparttable}
\end{center}
\vspace{0.35em}

\begin{center}
\tiny
\captionof{table}{\textbf{Design properties of \method{} relative to the closest methods.} Positioning,
\emph{not} a performance ranking; \method{} does not outperform the format-matched control
(main paper). \yes{}/\no{} mark whether a property holds; ``Selection
layer'' is where a method acts (selection vs.\ generation vs.\ confidence- or diagnosis-only);
``Generalizes maj.\ vote'' marks whether it contains majority voting as a special case.}
\label{tab:novelty}
\renewcommand{\arraystretch}{1.12}
\setlength{\tabcolsep}{8pt}
\resizebox{\textwidth}{!}{%
\begin{tabular}{@{}l ccccc@{}}
\toprule
\bfseries Method & \makecell{\bfseries Label-\\\bfseries free} & \makecell{\bfseries Train-\\\bfseries free}
 & \makecell{\bfseries Selection\\\bfseries layer} & \makecell{\bfseries General\\\bfseries VQA}
 & \makecell{\bfseries Generalizes\\\bfseries maj.\ vote} \\
\midrule
Self-consistency \citep{wang2023selfconsistency} & \yes & \yes & \yes & \yes & \emph{is MV} \\
Aha-Moment \citep{ahamoment2026}                  & \yes & \yes & \makecell{diag.\\only} & \yes & \textcolor{black!45}{n/a} \\
VALOR \citep{valor2026}                           & \yes & \no  & \makecell{gen.\\(RL)}  & \yes & \no \\
VTool-R1 \citep{vtoolr1_2026}                     & \no  & \no  & \makecell{gen.\\(RL)}  & \yes & \no \\
Zoom-Consist.\ \citep{zoomconsistency2026}        & \yes & \yes & \makecell{conf.\\only} & \no~{\scriptsize(GUI)} & \no \\
CRG \citep{crg2024}                               & \no & \yes & \yes & \yes & \no \\
See-not-Believe \citep{seeingbelieving2025}       & \yes & \yes & \makecell{diag.\\only} & \yes & \no \\
\midrule
\bfseries \method{} (ours) & \yes & \yes & \textbf{\textcolor{good}{selection}} & \yes & \yes \\
\bottomrule
\end{tabular}%
}
\end{center}
\vspace{0.35em}

\begin{center}
\footnotesize
\captionof{table}{\textbf{SensAblations perturbation-family ablations} (seed-mean over $\{0,23,42\}$ on the same
Qwen dumps as the main-paper results table; offline re-selection with \texttt{union} pool,
vote weight $1$, $\lambda{=}2$). Leave-one-family-out drops a single family from the full set;
family-only keeps just one. crop${=}$\{center, saliency\}, mask${=}$\{background\},
photo${=}$\{photometric\}, geom${=}$\{rotation, rescale\}. TextVQA is crop/geometry-sensitive;
other benchmarks are nearly flat across families.}
\label{tab:e5-family}
\renewcommand{\arraystretch}{1.08}
\resizebox{\textwidth}{!}{%
\begin{tabular}{@{}l *{5}{S[table-format=2.1]} *{4}{S[table-format=2.1]}@{}}
\toprule
 & \multicolumn{5}{c}{\bfseries Leave-one-family-out (\%)} &
   \multicolumn{4}{c}{\bfseries Family-only (\%)} \\
\cmidrule(lr){2-6}\cmidrule(lr){7-10}
\multicolumn{1}{@{}l}{\bfseries Benchmark ($N$)} &
\headrowc{Full} & \headrowc{$-$crop} & \headrowc{$-$mask} & \headrowc{$-$photo} & \headrowc{$-$geom} &
\headrowc{crop} & \headrowc{mask} & \headrowc{photo} & \headrowc{geom} \\
\midrule
TextVQA (300)    & 86.2 & 76.6 & 82.4 & 83.6 & 80.8 & 71.1 & 60.2 & 59.0 & 64.1 \\
MATH-V (299)     & 25.2 & 25.4 & 26.0 & 26.1 & 26.1 & 26.5 & 26.8 & 24.8 & 26.2 \\
MMMU (900)       & 50.3 & 50.0 & 50.6 & 50.0 & 50.2 & 50.3 & 49.4 & 49.7 & 50.1 \\
ViLP-Score (600) & 52.3 & 52.4 & 53.4 & 52.3 & 52.2 & 53.3 & 51.7 & 53.3 & 53.3 \\
\bottomrule
\end{tabular}%
}
\end{center}
\vspace{0.35em}

\begin{center}
\footnotesize
\captionof{table}{\textbf{SensAblations hyperparameter and compute sweeps} (accuracy \%; seed-mean over
$\{0,23,42\}$, offline re-selection on the same Qwen dumps as the main-paper results table).
$\lambda{=}0$ recovers majority voting; the operating point is $\lambda{=}2$. The $K$ and $M$
sweeps fix $\lambda{=}2$: $K$ truncates samples per view, $M$ keeps the first $M$ perturbation
views. TextVQA improves monotonically with $M$ while $K$ has little effect; MATH-V peaks near
$\lambda{\approx}1$; MMMU is flat; ViLP's best points are near $\lambda{=}0$ / small $M$
(within ${\approx}1.5$\,pp of the operating point). Row maxima are shaded.}
\label{tab:e5-sweep}
\renewcommand{\arraystretch}{1.08}
\setlength{\tabcolsep}{2.8pt}
\resizebox{\textwidth}{!}{%
\begin{tabular}{@{}l *{6}{S[table-format=2.1]} *{4}{S[table-format=2.1]} *{6}{S[table-format=2.1]}@{}}
\toprule
 & \multicolumn{6}{c}{\bfseries $\lambda$ sweep (vw${=}1$)} &
   \multicolumn{4}{c}{\bfseries $K$ sweep} &
   \multicolumn{6}{c}{\bfseries $M$ sweep} \\
\cmidrule(lr){2-7}\cmidrule(lr){8-11}\cmidrule(lr){12-17}
\multicolumn{1}{@{}l}{\bfseries Bench.} &
{0} & {0.5} & {1} & {\bfseries 2} & {3} & {4} &
{1} & {2} & {3} & {4} &
{1} & {2} & {3} & {4} & {5} & {6} \\
\midrule
TextVQA & 54.4 & 65.2 & 75.2 & 86.2 & 88.7 & \cellcolor{hlcol}89.0
        & 85.9 & 85.8 & 86.1 & 86.2
        & 62.7 & 71.1 & 77.1 & 80.8 & 83.2 & 86.2 \\
MATH-V  & 24.2 & 25.6 & 26.0 & 25.2 & 24.6 & 24.3
        & \cellcolor{hlcol}26.8 & 25.7 & 25.1 & 25.2
        & 26.6 & 26.5 & 26.7 & 26.1 & 25.6 & 25.2 \\
MMMU    & 49.5 & 49.6 & 50.0 & 50.3 & 50.4 & 50.2
        & \cellcolor{hlcol}50.5 & 50.4 & 50.1 & 50.3
        & 50.1 & 50.3 & 50.0 & 50.2 & 50.2 & 50.3 \\
ViLP-S  & 53.2 & 53.1 & 52.9 & 52.3 & 52.0 & 52.0
        & 52.5 & 52.2 & 52.1 & 52.3
        & \cellcolor{hlcol}53.6 & 53.3 & 52.2 & 52.2 & 52.2 & 52.3 \\
\bottomrule
\end{tabular}%
}
\end{center}
\vspace{0.35em}

\begin{center}
\footnotesize
\captionof{table}{\textbf{SensAblations label-free $\lambda$ selection} via perturbation self-agreement
(seed-mean over $\{0,23,42\}$, same dumps as the main-paper results table). For each example
the $K$ samples of every perturbation are split into halves, selection is run on each, and
agreement of the chosen answers is recorded (dataset means). The rule picks the largest
$\lambda$ whose split-half agreement is ${\geq}80\%$; picking the smallest trivially yields
$\lambda{=}0$. The label-free choice $\hat\lambda$ matches the label-aware oracle
$\lambda^\star$ on TextVQA; on ViLP-Score the oracle prefers $\lambda{=}0$ (MV) while the
no-label rule still selects $\hat\lambda{=}4$.}
\label{tab:e5-lambda-select}
\renewcommand{\arraystretch}{1.08}
\resizebox{\textwidth}{!}{%
\begin{tabular}{@{}l *{6}{S[table-format=3.1]} c S[table-format=2.1] c S[table-format=2.1]@{}}
\toprule
 & \multicolumn{6}{c}{\bfseries Split-half self-agreement (\%)} &
   \multicolumn{2}{c}{\bfseries No-label rule} &
   \multicolumn{2}{c}{\bfseries Oracle} \\
\cmidrule(lr){2-7}\cmidrule(lr){8-9}\cmidrule(lr){10-11}
\multicolumn{1}{@{}l}{\bfseries Benchmark} &
\headrowc{$\lambda{=}0$} & \headrowc{0.5} & \headrowc{1} & \headrowc{2} & \headrowc{3} & \headrowc{4} &
\headrowc{$\hat\lambda$} & \headrowc{Acc} & \headrowc{$\lambda^\star$} & \headrowc{Acc} \\
\midrule
TextVQA    & 100.0 & 97.3 & 95.3 & 94.8 & 93.2 & 92.4 & {4} & 89.0 & {4} & 89.0 \\
MATH-V     & 100.0 & 87.3 & 79.4 & 68.4 & 64.2 & 62.2 & {0.5} & 25.6 & {1} & 26.0 \\
MMMU       & 100.0 & 95.2 & 92.3 & 88.0 & 85.1 & 84.0 & {4} & 50.2 & {3} & 50.4 \\
ViLP-Score & 100.0 & 98.7 & 97.8 & 96.0 & 95.1 & 94.1 & {4} & 52.0 & {0} & 53.2 \\
\bottomrule
\end{tabular}%
}
\end{center}
\vspace{0.35em}

\begin{center}
\footnotesize
\captionof{table}{\textbf{RoutingTest per-instance correlation between the StabilityGap and the per-example
gain} $(\mathbf{1}[\method{}]-\mathbf{1}[\mathrm{MatchedCtrl}])$, pooled over seeds $\{0,23,42\}$.
Near-zero $r$ (except a small negative association on MATH-V) shows the gap does not
reliably predict when \method{} beats the format-matched control.}
\label{tab:e6-corr-e2}
\renewcommand{\arraystretch}{1.08}
\begin{tabular}{@{}l S[table-format=3.0] S[print-implicit-plus,table-format=+1.3] S[table-format=1.3] S[print-implicit-plus,table-format=+1.3] S[print-implicit-plus,table-format=+2.1]@{}}
\toprule
\multicolumn{1}{@{}l}{\bfseries Benchmark} & \headrowc{$N$} &
\headrowc{Pearson $r$} & \headrowc{$p$} & \headrowc{Spearman $\rho$} & \headrowc{Mean gain (pp)} \\
\midrule
TextVQA & 900 & -0.041 & 0.22  & -0.034 & -1.3 \\
MATH-V  & 896 & -0.111 & 0.00088 & -0.119 & +0.7 \\
MMMU    & 2700 & +0.010 & 0.61  & +0.005 & +0.1 \\
ViLP-P  & 1800 & +0.052 & 0.03 & +0.031 & -1.4 \\
\bottomrule
\end{tabular}
\end{center}
\vspace{0.35em}

\begin{center}
\footnotesize
\captionof{table}{\textbf{RoutingTest mean gain of \method{} over MatchedCtrl (pp) by stability-gap quartile}, pooled
over seeds $\{0,23,42\}$. Q1--Q4 are increasing-gap quartiles (Q4 highest). Gain does not
increase monotonically with the gap.}
\label{tab:e6-quart-e2}
\renewcommand{\arraystretch}{1.08}
\begin{tabular}{@{}l S[print-implicit-plus,table-format=+1.1] S[print-implicit-plus,table-format=+1.1] S[print-implicit-plus,table-format=+1.1] S[print-implicit-plus,table-format=+1.1]@{}}
\toprule
\multicolumn{1}{@{}l}{\bfseries Benchmark} & \headrowc{Q1} & \headrowc{Q2} & \headrowc{Q3} & \headrowc{Q4} \\
\midrule
TextVQA & +1.8 & -1.8 & -4.9 & -0.4 \\
MATH-V  & +9.4 & -0.4 & -0.9 & -5.4 \\
MMMU    & +0.0 & +0.1 & -0.7 & +1.0 \\
ViLP-P  & +0.0 & -6.2 & +0.4 & +0.2 \\
\bottomrule
\end{tabular}
\end{center}
\vspace{0.35em}

\begin{center}
\footnotesize
\captionof{table}{\textbf{RoutingTest ViLP recoveries of \method{} relative to MatchedCtrl} (non-prior Score slots,
mean over seeds $\{0,23,42\}$). Of the grounded questions MatchedCtrl answers incorrectly, \method{}
recovers only a few; mechanism-targeted rescues remain rare. Prior-aligned error breakdown
is omitted (prior-option labels are not stored in these dumps).}
\label{tab:e6-vilp-e2}
\renewcommand{\arraystretch}{1.08}
\begin{tabular}{@{}l r@{}}
\toprule
\bfseries Quantity & \multicolumn{1}{r}{\bfseries Value} \\
\midrule
Score slots ($N$; pooled over 3 seeds)           & 1800 \\
MatchedCtrl wrong on grounded                             & 834 \\
\quad \method{} flips MatchedCtrl-wrong $\to$ correct     & 13 (1.6\%) \\
\bottomrule
\end{tabular}
\end{center}
\vspace{0.35em}

\putbib
\end{bibunit}

\end{document}